%% file: ijcai22-multiauthor.tex
\documentclass{article}
\usepackage{ijcai22}
\usepackage{times}

\usepackage{soul}
\usepackage{url}
\usepackage[hidelinks]{hyperref}
\usepackage[utf8]{inputenc}
\usepackage[small]{caption}
\usepackage{graphicx}
\usepackage{amsmath}
\usepackage{amsfonts}
\usepackage{booktabs}
\usepackage{xcolor}
\usepackage{subcaption}
\usepackage{graphicx}

\title{Test-time Fourier Style Calibration for Domain Generalization}

\author{
Xingchen Zhao$^1$
\and
Chang Liu$^1$\and
Anthony Sicilia$^2$\and
Seong Jae Hwang$^2$ \And
Yun Fu $^1$
\affiliations
$^1$Northeastern University\\
$^2$University of Pittsburgh
\emails
\{zhao.xingc, liu.chang6\}@northeastern.edu,
\{anthonysicilia, sjh95\}@pitt.edu,
yunfu@ece.neu.edu
}
\begin{document}

\maketitle

\input{1_abs.tex}
\input{2_intro.tex}
\input{3_related_work}
\input{4_method}
\input{5_exp}

\input{6_conclusion}
\bibliographystyle{named}
\bibliography{ijcai22}
\end{document}

%% file: 1_abs.tex
\begin{abstract}
The topic of generalizing machine learning models learned on a collection of source domains to unknown target domains is challenging. While many domain generalization (DG) methods have achieved promising results, they primarily rely on the source domains at train-time without manipulating the target domains at test-time. Thus, it is still possible that those methods can overfit to source domains and perform poorly on target domains. Driven by the observation that domains are strongly related to styles, we argue that reducing the gap between source and target styles can boost models' generalizability. To solve the dilemma of having no access to the target domain during training, we introduce \textbf{T}est-time \textbf{F}ourier Style \textbf{Cal}ibration (TF-Cal) for calibrating the target domain style on the fly during testing. To access styles, we utilize Fourier transformation to decompose features into amplitude (style) features and phase (semantic) features. Furthermore, we present an effective technique to \textbf{A}ugment \textbf{A}mplitude \textbf{F}eatures (AAF) to complement TF-Cal. Extensive experiments on several popular DG benchmarks and a segmentation dataset for medical images demonstrate that our method outperforms state-of-the-art methods.
\end{abstract}

%% file: 2_intro.tex
\section{Introduction}
Over the past few years, deep convolutional neural networks (CNNs) have been shown to perform excellently in a variety of computer vision applications. However, CNNs have demonstrated a drastic performance drop when both the training (source domain) and testing (target domain) data come from different distributions. This incapability causes numerous problems in real-world ﬁelds such as autonomous driving and medical imaging analysis. As a consequence, domain adaptation (DA) has been extensively researched as a mechanism of transferring knowledge from source domains to labeled/unlabeled target domains. Yet, DA approaches rely on known target domains at training time, which fails to generalize well to unknown testing target domains.
% \begin{figure}
%     \centering
%     \includegraphics[width=1\columnwidth]{figs/Fig1.pdf}
%     \vspace{-15pt}
%     \caption{Our approach uses the estimated source prototype to calibrate the target domain at test-time. It is important to distinguish between the original source (solid circles) and the augmented source distribution (dashed circles). The augmentation can further reduce the gap between calibrated target and source domains.}
%     \label{fig:fig1}
%     \vspace{-13pt}
% \end{figure}
Domain generalization (DG) has lately attracted much attention as a more challenging setting for enhancing CNNs' generalizability to unknown testing data. Specifically, DG is to train models that are resistant to distribution shifts between domains (i.e., domain shifts) so that models can generalize well when evaluated on any previously unseen target domain, provided that several source domains are available for training. To address the problem, various methods were proposed and achieved promising results, including adversarial learning \cite{Zhou2020DeepDI,Sicilia2021DomainAN}, meta learning \cite{Li2018LearningTG,Balaji2018MetaRegTD}, and data augmentation \cite{Xu2021AFF,Zhou2021DomainGW}. Among these, approaches based on data augmentation have gained popularity due to their simplicity and effectiveness, which can expose models to a wide variety of source domains to learn generalizable representation. However, most existing data augmentation approaches cannot perform well on test data that are out of the augmented source data distribution. This begs the question of how we can reduce domain gaps when the augmented source domains are insufficient to cover the target domain.

In this paper, we propose a novel yet simple method called Test-time Fourier Style Calibration (TF-Cal) to calibrate the target style features at inference-time. Our method is driven by the assumption that the lack of generalizability is due to the difference in styles of images (e.g., sketches, cartoon, photo) between the source and target \cite{Zhou2021DomainGW}. Based on this idea, we employ Fourier transformation to decompose latent features from early CNN layers into amplitude features and phase features. The amplitude encodes the intensities (style information), while the phase encodes the spatial positions (semantic information) \cite{Piotrowski1982ADO}. To achieve test-time style calibration, we propose to convexly calibrate the target amplitude features using the source amplitude prototype estimated at train-time. Hence we narrow the style gap between the target and source domains on the fly when evaluating, while retaining the key semantic feature for classification. We additionally show that increasing the diversity of features can complement TF-Cal to further boost models' generalizability based on the convex hull theory proposed by \cite{Albuquerque2019GeneralizingTU}. Inspired by \cite{Xu2021AFF} which augments amplitude in image space, we propose augmenting amplitude in feature space (AAF) that diversifies higher-level styles, while preserving semantics. Then we have the \textbf{T}est-time \textbf{A}ugmentated \textbf{F}ourier Style \textbf{C}alibration (TAF-Cal) by combining TF-Cal and AAF. Moreover, our algorithm does not require domain labels, allowing for more data collection flexibility.

The proposed work is evaluated on multiple DG benchmarks with different backbones on image classification and medical image segmentation. Through extensive experiments, we demonstrate that our method significantly improves the generalizability of CNNs and outperforms multiple state-of-the-art methods. We also show that TF-Cal is useful in conjunction with augmentation techniques. Finally, we present qualitative analysis and quantitative ablation studies to verify the effectiveness of our method.

%% file: 3_related_work.tex
\section{Related Work}
\subsection{Domain Generalization}
DG works in recent years can be roughly categorized into: domain-invariant learning, meta-learning and data augmentation approaches.

\textbf{Domain-invariant learning} aims to map input data to features that are invariant to source domain shift in order to be resilient to unseen target domain shift. The domain adversarial neural network (DANN) \cite{Ganin2016DomainAdversarialTO}, which was originally developed for DA, consists of a domain classifier trained to discriminate between the source and target distributions, and a feature extractor trained to learn features that are not only beneficial for the task (e.g., classifcation) but also capable of confusing the domain classifier. \cite{Albuquerque2019GeneralizingTU,Sicilia2021DomainAN} leverage the DANN to align the representation of source domains to learn the domain invariance in DG. \cite{Li2018DomainGW} propose to use Maximum Mean Discrepancy with adversarial autoencoder to align domain distributions, mapping the aligned distribution to a prior distribution adversarially. Another line of works focus on minimizing contrastive loss to learn domain-invariant representation. MASF \cite{Dou2019DomainGV} minimizes contrastive losses with positive pair of same class and negative pair of different class.

\textbf{Meta-learning} based approach MLDG \cite{Li2018LearningTG} simulates domain shift by dividing data from multiple sources into meta-train and meta-test sets. \cite{Balaji2018MetaRegTD} propose a meta-regularizer learned to be robust to domain shift by meta-learning. \cite{Dou2019DomainGV}   regularize semantic structure of features while optimizing models by meta-learning. While meta-learning methods can simulate the domain shift during training by dividing source domains to meta-train and meta-test sets, it is still likely that the simulated domain shift cannot cover the target domain shift.

\textbf{Data augmentation} method is most related to our work, attempting to augment the training images or features to enhance the generalization performance on unseen data. \cite{Zhou2021DomainGW} probabilistically mix the instance statistics of training features. \cite{Xu2021AFF} propose to augment amplitude of images with a regularization to form consistency between original and augmented images. To generate diverse novel training images, \cite{Zhou2020DeepDI} design an image generation network with domain-adversarial training plus optimal transport based metrics.  A concurrent work \cite{Zhang2021MoreIB} has a similar motivation as ours to augment/modify target data during evaluation, developing a multi-view regualizeried meta-learning that encourages test-time augmentation with multi-view.  Although these methods achieve favorable results on unseen targets, they might fail when the augmented source cannot cover the unseen target.  Our method mitigates this issue by calibrating the target with a source prototype, allowing the calibrated target to be covered by the source.

%% file: 4_method.tex
% \clearpage 
% \newpage
\begin{figure*}
    \centering
    \hspace*{-0.30cm}\includegraphics[]{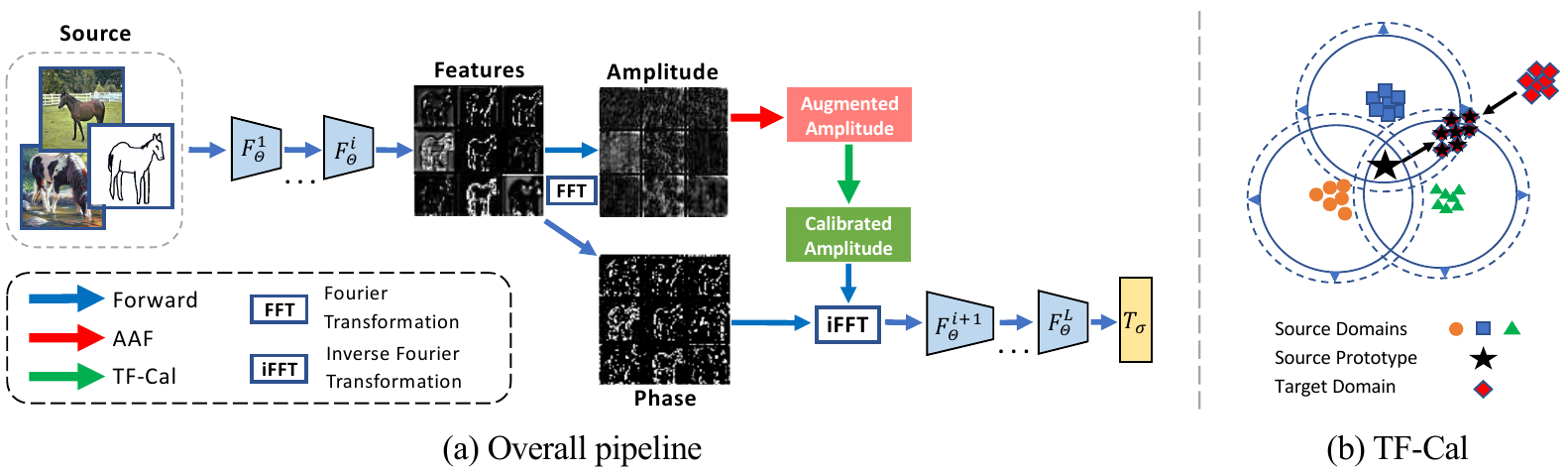}
    \vspace{-10pt}
    \caption{Our method's overview. The overall pipeline is illustrated in (a). The images are fed into the feature extractor, and the early layer' features are then decomposed into amplitude and phase features. We then augment the amplitude (AAF). Following that, the model uses TF-Cal to calibrate the amplitude features. Next, the modified amplitude with the unmodified phase are inversed into the spatial domain. Note that the AAF is only used at train-time. Remark: For visualization purpose, we reconstruct the amplitude and phase with a constant to the spatial domain. (b) illustrates TF-Cal, where solid circles represent the original source and dashed circles represent the augmented source.}
    \label{fig:fig2}
    \vspace{-5pt}
\end{figure*}
\section{Methodology}
\subsection{Problem Definition}
We define $\mathcal{X}$ and $\mathcal{Y}$ as the input and label space. Given $K$ source domains $\mathcal{D}_{s}=\left\{D^{1}_{s}, D^{2}_{s}, \ldots, D^{K}_{s}\right\} (1 \leq i \leq K)$ that are available at train-time, we indicate that each $D^{i}_{s}$ contains $N$ input and label pairs $(x_{j}, y_{j})_{j=1}^{N_{i}}$ where ${x} \in \mathcal{X} , y \in \mathcal{Y}$ and $N_{i}$ is the number of inputs in the source domain $D^{i}_{s}$. The goal of DG is to train a function learned from data of $\mathcal{D}_{s}$ to perform well on an unseen target domain $\mathcal{D}_{t}$ which share the same label space $\mathcal{Y}$ as $\mathcal{D}_{s}$. The function can be defined as $R_{\theta,\sigma}=T_{\sigma} \circ F_{\theta}$, where $T_{\sigma}$ is a task-specific network parameterized by $\sigma$ and $F_{\theta}=\left\{F^{1}_{\theta}, F^{2}_{\theta}, \ldots, F^{L}_{\theta}\right\} (1 \leq i \leq L)$ which is a feature encoder with $L$ layers parameterized by $\theta$. 

\subsection{Background}
To reduce the domain shift, we need to first understand the difference between two distributions which are measured by $\mathcal{H}$-divergence proposed by \cite{BenDavid2009ATO}: 
\begin{equation}
    d_{\mathcal{H}}(\mathcal{D}_{s}, \mathcal{D}_{t})=2 \sup _{h \in \mathcal{H}}\left|\operatorname{Pr}_{x\sim{\mathcal{D}_{s}}}[h(x)=1]-\operatorname{Pr}_{x\sim{\mathcal{D}_{t}}}[h(x)=1]\right|
\end{equation}
where classifier $h: \mathcal{X} \rightarrow\{0,1\}$.
% In this case, $\mathcal{H}$-divergence can be estimated by how well a binary classifier can distinguish between domains.
Following that, \cite{Albuquerque2019GeneralizingTU} defines the convex hull $\Lambda_{s}$ of $\mathcal{D}_{s}$ that is a set of mixture of source domain distributions:
\begin{equation}
\Lambda_{s}=\left\{\sum_{i=1}^{K} \pi_{i} \mathcal{D}_{s}^{i} \mid \pi \in \Delta_{K-1}\right\}
\end{equation}
where the $\pi$ represents non-negative coefficient in the $(K-1)$-dimensional simplex $\Delta_{K-1}$. Then, an ideal case $\Bar{\mathcal{D}}_{t} \in \Lambda_{s}$ is assumed that the ideal target $\Bar{\mathcal{D}}_{t}$ lies in the source domain convex hull $\Lambda_{s}$. Under this assumption, the risk $\epsilon_{t}(h)$ on the target domain is upper-bounded \cite{Albuquerque2019GeneralizingTU} by:
\begin{equation}
\epsilon_{t}(h) \leq  \sum_{i=1}^{K} \pi_{i} \epsilon^{i}_{s}(h) + \gamma + \zeta
\label{eq:3}
\end{equation}
where, on the right hand side, the first term corresponds to the risks over all source domains, and the second term $\gamma=d_{\mathcal{H} \Delta \mathcal{H}}(\Bar{\mathcal{D}}_{t}, D_{t})$ is the $\mathcal{H}$-divergence between the ideal target $\Bar{\mathcal{D}}_{t}$ and the real target $D_{t}$, and the third term $\zeta=\sup _{D_{s}^{\prime}, D_{s}^{\prime \prime} \in \Lambda_{s}} d_{\mathcal{H} \Delta \mathcal{H}}\left(D_{s}^{\prime}, D_{s}^{\prime \prime}\right)$ which is the largest $\mathcal{H}$-divergence between any pair of source domains. $\mathcal{H} \Delta \mathcal{H}$ corresponds to $\left\{h(x) \oplus h^{\prime}(x) \mid h, h^{\prime} \in \mathcal{H}\right\}$. The first term can be minimized by empirical risk minimization (ERM), and the second term is hard to minimized due to having no access to the target domain at training, and the third term can be minimized by removing the source domain-discrimintive information which is the style information in the context of this paper. According to \cite{Cha2021DomainGN}, ERM is a strong baseline to learn source domain invariance compared to other methods such as DANN \cite{Ganin2016DomainAdversarialTO}.

\subsection{Motivation}
As the Eq.\eqref{eq:3} states, a model can achieve acceptable performance on the target domain by using ERM to train on the source domains when the ideal target domain $\Bar{\mathcal{D}}_{t}$ is assumed to be covered by source convex hull $\Lambda_{s}$. However, in reality, this assumption typically does not hold. This is the reason why domain-adversarial based methods \cite{Zhou2020DeepDI,Albuquerque2019GeneralizingTU} may fail in DG as they only reduce source divergence. Therefore, augmenting images or features of source domains can possibly extend the $\Lambda_{s}$ (i.e., $\gamma \rightarrow 0$), which motivating a variety of works \cite{Zhou2021DomainGW,Sicilia2021DomainAN} to diversify training data. Still, as illustraed in Fig.\ref{fig:fig2}b, it is likely that the augmented source distributions (dashed circles) may not accurately reflect the previously unseen target domain (red rhombi). This difficulty arises mostly due to the absence of target data during training. Instead of extending $\Lambda_{s}$ arbitrarily, we aim to pull $D_{t}$ closer to $\Bar{D_{t}}$ with the help from source data.

Therefore, it motivates us to calibrate the target features at test-time based on what a model learned from source features. By doing so, we pull the target closer to the source convex hull $\Lambda_{s}$, and reduce the difference between $D_{t}$ and $\Bar{D_{t}}$. To further encourage the reduction in source-target divergence, we augment the source features to expand the convex hull.

\subsection{Method Overview}
Our method TF-Cal is to calibrate style features. To this end, we use Fourier transformation to decompose features into phase features that contain semantics and amplitude features that contain styles. Then we let a model to learn with a style calibrating function $\mathcal{C}$ (TF-Cal) for reducing the gap between the target domain and source domains prototype. We further introduce an augmenting function $\mathcal{A}$ (AAF) for expanding the source convex hull by diversifying style features. The model then is trained in a two-stage manner: in the early stage, we train the model with AAF to learn a great diversity of style features, and in the late stage, we train the model with both AAF and TF-Cal (i.e., TAF-Cal) that uses the estimated source prototype from diverse styles. During testing, we leverage the learned $\mathcal{C}$ to calibrate target styles. Our method is easy to implement as a plug-and-play module without adding any additional network. The overall pipeline is illustrated in Fig.\ref{fig:fig2}a and TF-Cal is illustrated in Fig.\ref{fig:fig2}b, and the detail of the method will be explained as follows.

\subsection{Fourier Transformation of Features}
Given a latent feature tensor $\mathbf{z}^{i}=F^{i}(\mathbf{z}^{i-1})$ with $\mathbf{z}^{i} \in \mathbb{R}^{N \times C \times H \times W}$ computed from $i$-th layer of the feature extractor, we want to extract style information from the feature. In our implementation, we choose the early layers since they can extract style information well \cite{Zhou2021DomainGW}. Fourier transformation $\mathcal{F}(\mathbf{z}^{i})$, which can be implemented using Fast Fourier transformation (FFT) \cite{Nussbaumer1982TheFF}, is able to decompose the feature $\mathbf{z}^{i}$ into the phase spectrum $\mathcal{PH}(\mathbf{z}^{i})$ encoding the spatial information and amplitude spectrum $\mathcal{AM}(\mathbf{z}^{i})$ encoding the intensity information. As illustrated in Fig.\ref{fig:fig2}, the amplitude component of the horse features depicts the intensity of texture, while phase component preserves the shape of object. Thus, it is natural to associate the amplitude with styles and the phase with semantics. Since our method only modifies the amplitude, the semantic information in the phase will not be affected.

The Fourier transformation $\mathcal{F}(\mathbf{z}^{i})$ is done on each channel $C$ of $\mathbf{z}^{i}$:
\begin{equation}
    \mathcal{F}(\mathbf{z}^{i})(m, n)=\sum_{h=0}^{H-1} \sum_{w=0}^{W-1} \mathbf{z}^{i} (h, w) e^{-j 2 \pi}\left(m\frac{h}{H} + n\frac{w}{W} \right)
\end{equation}
where $\mathcal{F}(\mathbf{z}^{i})(m, n)$ represents the value of the features in the frequency space corresponding to the coordinates $m$ and $n$. With $\mathcal{F}$, we can retrieve the amplitude and phase features with real $\operatorname{Re}$ and imaginary $\operatorname{Im}$ parts:
\vspace{0.5pt}
\begin{equation}
\begin{aligned}
&\mathcal{PH}(\mathbf{z}^{i})=\arctan \frac{\operatorname{Im}(\mathcal{F}(\mathbf{z}^{i})(m, n))}{\operatorname{Re}(\mathcal{F}(\mathbf{z}^{i})(m, n))}\\
&\mathcal{AM}(\mathbf{z}^{i})=\sqrt{\operatorname{Im}(\mathcal{F}(\mathbf{z}^{i})(m, n))^{2}+\operatorname{Re}(\mathcal{F}(\mathbf{z}^{i})(m, n))^{2}}
\end{aligned}
\end{equation}
To reconstruct phase and amplitude spectrum features back into the spatial feature space, we can use the inverse Fourier Transformation $\mathcal{F}^{-1}(\mathcal{AM}(\mathbf{z}^{i}),\mathcal{PH}(\mathbf{z}^{i}))=\mathbf{z}^{i}$.

\subsection{Test-time Fourier Style Calibration}
% \vspace{2.5pt}
\subsubsection*{Estimating Source Prototype}
The first step in our algorithm TF-Cal is to estimate the source domain prototype $\mathbf{Pt}_{s} \in \mathbb{R}^{1 \times C \times H \times W}$ that will be used for the calibrating function $\mathcal{C}$.
The $\mathbf{Pt}_{s}$ can be estimated by taking the average over the amplitude features of all data points $N_{D}$ from source domains along the batch dimension:
\begin{equation}
    \mathbf{Pt}_{s}=\frac{1}{N_{D}}\sum_{n=0}^{N_{D}-1} \mathcal{AM}(\mathbf{z}^{i}_{s})
\end{equation}
Since the $\mathbf{Pt}_{s}$ is estimated via all training samples, it will be a proper representation of the centroid of source styles. We leverage $\mathbf{Pt}_{s}$ as the anchor to pull a style closer to the source convex hull $\Lambda_{s}$.
In practice, we store the $\mathcal{AM}(\mathbf{z}^{i}_{s})$ from each mini-batch in a memory bank, and compute the mean of all amplitude features from the bank at the end of one epoch. We will verify calibrating style with source prototype instead of a random prototype in the ablation studies section. 
\subsubsection*{Training with the Style Calibrating Function}
The TF-Cal is proposed to reduce the divergence between source and target styles during testing. We need to first train a model with a calibrating function $\mathcal{C}$ with the input of the source prototype $\mathbf{Pt}_{s}$ and source amplitude features $\mathcal{AM}(\mathbf{z}^{i}_{s})$. Therefore, we let the model's parameters be associated with the source prototype that is the common information shared by the target and source styles. The $\mathcal{C}$ is used to train the model by linearly interpolating between $\mathcal{AM}(\mathbf{z}^{i}_{s})$ and the $\mathbf{Pt}_{s}$:
\begin{equation}
    \mathcal{C}(\mathbf{Pt}_{s},\mathcal{AM}(\mathbf{z}^{i}_{s}))=\eta\mathbf{Pt}_{s}+(1-\eta)\mathcal{AM}(\mathbf{z}^{i}_{s})
\label{eq:7}
\end{equation}
where $\eta$ is the hyperparameter to control the strength of calibration. 
The calibrating function $\mathcal{C}$ is activated with a probability $p_{cal}$ during training. Before the training is finished, we store the source prototype $\mathbf{Pt}_{s}$ from the final epoch for the usage of calibrating target style at test-time. 
\subsubsection*{Calibrating the Target Style on the Fly}
During the evaluation, we leverage the $\mathcal{C}$ function to calibrate the target style $\mathcal{AM}(\mathbf{z}^{i}_{t})$ with the source prototype $\mathbf{Pt}_{s}$:
\begin{equation}
    \mathcal{C}(\mathbf{Pt}_{s},\mathcal{AM}(\mathbf{z}^{i}_{t}))=\tau\mathbf{Pt}_{s}+(1-\tau)\mathcal{AM}(\mathbf{z}^{i}_{t})
\label{eq:8}
\end{equation}
where the $\tau$ is the hyperparamter to control the strength of calibration during testing. The reason for controlling $\tau$ is that, due to CNNs' inductive biases toward styles \cite{Geirhos2019ImageNettrainedCA}, target amplitude may contain useful style information for classification. We set the calibration probability $p_{cal}$ to 1 during testing. Our method requires only one test sample $\mathbf{z}^{i}_{t} \in \mathbb{R}^{1 \times C \times H \times W} $ for calibration which is realistic in practical applications. Through calibration, the target style is pulled closer to the source style, helping the model to achieve a similar performance on target data as it does on source data. The advantage of our method is that we let the target and source styles share the $\mathbf{Pt}_{s}$ that is controllable and at hand. Thus, we can offset the source-target divergence with the guidance from the source prototype at test-time. Instead, other DG methods only rely on the source data, hoping to cover the target distribution or learn domain-invariance that overfits to source domains during training. 

% More related works on test-time adaptation require to adapt models' parameters to the target which unnecessarily increase the inference time.
\subsection{Augmenting Amplitude Features}
As we can see from Eq.\eqref{eq:7} and Eq.\eqref{eq:8}, the common part of the input is the $\mathbf{Pt}_{s}$ and the different part is the amplitude features $\mathcal{AM}(\mathbf{z}^{i}_{t})$ and $\mathcal{AM}(\mathbf{z}^{i}_{s})$. Therefore, to further boost the generalizability with TF-Cal, we want $\mathcal{AM}(\mathbf{z}^{i}_{t})$ and $\mathcal{AM}(\mathbf{z}^{i}_{s})$ to be as similar as possible. An effective way is to augment amplitude in feature space (AAF) hoping to expand the source convex hull $\Lambda_{s}$ to decrease the source-target divergence. Therefore, based on the strategy of mixing and swapping amplitude of images \cite{Xu2021AFF}, we propose to mix and swap amplitude feature pairs ($\mathcal{AM}(\mathbf{z}^{i}_{s}),\mathcal{AM}(\mathbf{z}^{i'}_{s})$) randomly from any source domains:
\begin{equation}
    \mathcal{A}(\mathcal{AM}(\mathbf{z}^{i}_{s}),\mathcal{AM}(\mathbf{z}^{i'}_{s}))=\delta\mathcal{AM}(\mathbf{z}^{i}_{s})+(1-\delta)\mathcal{AM}(\mathbf{z}^{i'}_{s})
\end{equation}
where $\delta \sim \operatorname{Beta}(\alpha, \alpha)$. For swapping amplitude between two samples, we set $\delta$ to 0. Compared to \cite{Xu2021AFF} that augment styles in the image space, AAF utilizes higher level style information from features with much more channels, which provides additional statistics for diversifying styles. In our implementation, we choose amplitude features from early layer of feature extractors since they encode style information about the inputs. Our results in the ablation studies demonstrate that augmenting amplitude on feature is better than on image, which empirically verify that AAF helps TF-Cal. 

% Albuquerque consider a situation where the target domain distribution lies within the convex hull of source domain distributions. The convex hull is defined as the convex combination. In this case, reduce the risk over source domain 

%  The amplitude encodes the intensities (carrying style information), while the phase encodes the spatial positions (carrying semantic information) that describes how those intensities lie in juxtaposition to one another.
 
%  At train-time, we let the model to learn a calibration function which maps a distribution to a calibrated distribution. 

% In reality, domain adaption bound is not suitable for domain generalization since we cannot access target domain during training. However, with the test-time calibration, it is possible to access target and then reduce the divergence on the fly.

%The previous fourier based DG method may not be effective to generalize, because the variability it learns during training is not represented in source-target shift. Unlike them, we let a model to learn a common style representation between source and target, which is the source style prototype. Thus, during testing, the source-target shift is reduced by calibrating the target data with the common source prototype.

%% file: 5_exp.tex
\section{Experiments}
\subsection{Datasets and Settings}
Our method is evaluated on three DG benchmarks: (1) \textbf{PACS} \cite{Li2017DeeperBA} is an object recognition dataset that contains 4 different domains (i.e., \textbf{P}hoto, \textbf{A}rt-Painting, \textbf{C}artoon, \textbf{S}ketch), including 9,991 images and 7 classes. We follow the train-val split provided by \cite{Li2017DeeperBA}. The domain shift is primarily contributed by the style difference. (2) \textbf{Office-Home} \cite{Venkateswara2017DeepHN} is an object recognition dataset that contains 4 different domains (i.e., Art, Clipart, Product, Real-World), including 15,550 images and 65 classes. The dataset is split randomly into 90\%
training set and 10\% validation set. The domain shift is primarily contributed by the style and viewpoint differences, with less diversity than PACS. (3) \textbf{MICCAI WMH Challenge}  \cite{Kuijf2019StandardizedAO} is a medical image segmentation dataset for \textbf{W}hite \textbf{M}atter \textbf{H}yperintensity that contains 3 different domains (i.e., Utrecht, Amsterdam, Singapore), including 20 subjects in each domain. For each subject, there are 47 slices in Utrecht and Singapore, and 82 slices in Amsterdam. The domain shift is mainly due to scanner and protocol differences.

For evaluation, we use the leave-one-domain-out protocol. In other words, the model is trained on source domains, and is evaluated on a held-out target domain. For PACS and OfficeHome, we report the classification accuracy. For WMH Challenge, we report the Dice Similarity Coefficient (DSC) (higher is better). All results are averaged over 3 runs with different random seeds. We use DeepAll as the ERM baseline with no DG mechanisms.

\subsection{Implementation Details}
(1) For PACS and OfficeHome, we employ the ResNet \cite{He2016DeepRL} pretrained on the ImageNet as the backbone. The networks are trained with the learning rate of 1e-3 for feature extractors, the learning rate of 1e-4 for classifiers, batch size of 64, and epochs of 50. We use the features from the second residual block as the input for AAF and TF-Cal. We adopt the same data augmentation as \cite{Matsuura2020DomainGU}. (2) For WMH Challenge, we use the U-Net \cite{Ronneberger2015UNetCN} trained from scratch as the backbone with the the learning rate of 2e-4, batch size of 30, and epochs of 300. We use the features from the second downsamling layer as the input for our method. We randomly augment each training slices by rotation, scaling, and shearing. All networks are optimized by SGD with weight decay of 5e-4, and all learning rates are decayed by 0.1 after 80\% of the epochs. For all experiments: For TF-Cal, we set the calibration strength $\eta$ and $\tau$ to 0.5, and set the $p_{cal}$ to 0.5. For AAF activated with a probability of 0.5, we set $\delta \sim \operatorname{Beta}(0.2, 0.2)$. The AAF is applied for all epochs, and TF-Cal is trained after 70\% of the epochs.

\subsection{Evaluation on PACS}
From the results in Table \ref{table:pacs_comparison}, our method outperforms the SOTA methods on both ResNet-18 and ResNet-50, improving the DeepAll by a large margin, which demonstrates the effectiveness of our method on unseen target domains. In comparison to the meta-learning methods MLDG \cite{Li2018LearningTG} and MASF \cite{Dou2019DomainGV}, as well as the domain-adversarial method MMLD \cite{Matsuura2020DomainGU}, our method clearly outperforms them significantly with simple feature transformations and no additional network. With the advantage of alleviating the overfit to source domains, our method also surpasses data augmentation methods DDAIG \cite{Zhou2020DeepDI} and MixStyle \cite{Zhou2021DomainGW}, as well as a more recent method, RSC \cite{Huang2020SelfChallengingIC}. FACT \cite{Xu2021AFF} is most related to our work in terms of Fourier transformation and augmentation. Still, our method outperforms FACT: (1) TF-Cal can calibrate target style at test time, which solves the problem of training with no knowledge of target domains. (2) AAF can encourage more diverse styles of source domains by providing additional signals in higher-level style information via features from early CNN layers. MVRML \cite{Zhang2021MoreIB} is relevant to our work in terms of test-time augmentation because it relies on augmenting target images with multiple views. Our method beats MVRML since we can calibrate target images with guidance from the source prototype.

\subsection{Evaluation on Office-Home}
As Table \ref{table:office_comparison} shows, our method has the best performance. For this dataset, we have similar observations and conclusions as the PACS. Our method shows improvement consistently over all target domains. Note that, Clipart has the most style difference compared to other domains. Our method demonstrates its ability to calibrate the target style to offset the source-target shifts, which improves the performance on Clipart by a large margin over the DeepAll.

\subsection{Evaluation on WMH Challenge}
We evaluate our method on the medical image segmentation task,  which is important for real-world applications and is often ignored by other DG works. Our method outperforms other related methods: (1) BigAug \cite{Zhang2020GeneralizingDL} applies heavy data augmentation. (2) UDA \cite{Li2020DomainAM} is an unsupervised domain adaptation method. (3) MixDANN \cite{Zhao2021RobustWM} is a method that combines mixup and DANN. We notice that the DeepAll fails to generalize to the Utrecht due to the significant scanner-difference. Our method significantly improves the overall performance of Utrecht by a considerable margin compared to previous methods.

\begin{table}[t!]
\resizebox{\columnwidth}{!}{%
\centering 
\footnotesize
\setlength{\tabcolsep}{4pt}
\renewcommand{\arraystretch}{1.10}
\begin{tabular}{l|cccc|c}
\specialrule{.1em}{.1em}{0.1em} 
\textbf{Method} & \textbf{Art} & \textbf{Cartoon} & \textbf{Photo} & \textbf{Sketch} & \textbf{Avg} \\ \hline
\multicolumn{6}{c}{\textbf{ResNet-18}}                                                                                                   \\ \hline
DeepAll       & 77.95        & 77.92            & 95.68           & 69.56          & 80.28                      \\
MLDG [2017] & 79.50        & 77.30            & 94.30           & 71.50          & 80.65                    \\
MASF [2019]         & 80.29        & 77.17            & 94.99           & 71.69          & 81.04                    \\
MMLD [2019]         & 81.28        & 77.16            & 96.09           & 72.29          & 81.83                    \\ 
DDAIG [2020]        & 84.20        & 78.10            & 95.30           & 74.70          & 83.08                       \\
MixStyle [2021]    & 84.10        & 78.80            & \textbf{96.10}           & 75.90          & 83.73                    \\ 
FACT [2021]          & 85.37        & 78.38            & 95.15           & 79.15          & 84.51                      \\
RSC [2020]          & 83.43        & 80.31            & 95.99           & 80.85          & 85.15                    \\
MVRML [2021]    & 84.59        & 79.22            & 95.38           & \textbf{83.55}          & 85.69                    \\
TAF-Cal (Ours)          & \textbf{85.70}        & \textbf{82.55}            & 96.07           & 82.60          & \textbf{86.73}                     \\
\hline\multicolumn{6}{c}{\textbf{ResNet-50}}                                                                                                    \\ \hline
DeepAll       & 84.91        & 78.58             & 97.78          & 73.89           & 83.79                     \\
MASF [2019]         & 82.89        & 80.49             & 95.01          & 72.29           & 82.67                   \\
MVRML [2021]         & 87.36        & 83.12             & 96.41          & 84.11           & 87.75                 \\
RSC [2021]           & 87.89        & 82.16             & \textbf{97.92}          & 83.35           & 87.83                 \\
FACT [2021]          & 89.63        & 81.77             & 96.75          & \textbf{84.46}           & 88.15                \\
TAF-Cal (Ours)          & \textbf{90.53}        & \textbf{84.91}             & 97.84          & 83.83           & \textbf{89.28}                 \\

\specialrule{.1em}{.1em}{0.1em} 

\end{tabular}
}
\caption{\label{table:pacs_comparison}Evaluation accuracy (\%) on PACS.}
\end{table}

\begin{table}[t!]
\resizebox{\columnwidth}{!}{%
\centering 
\footnotesize
\setlength{\tabcolsep}{4pt}
\renewcommand{\arraystretch}{1.10}
\begin{tabular}{l|cccc|c}
\specialrule{.1em}{.1em}{0.1em} 
\textbf{Method} & \textbf{Art} & \textbf{Clipart} & \textbf{Product} & \textbf{Real} & \textbf{Avg} \\ \hline                                           
DeepAll       & 59.93        & 48.80            & 74.13           & 75.42          & 64.57                      \\
RSC [2020]          & 58.42        & 47.90             & 71.63          & 74.54           & 63.12                \\
DDAIG [2020]     & 59.20          & 52.30           & 74.60            & 76.00          & 65.53                 \\
MixStyle [2021] & 58.70           & 53.40           & 74.20            & 75.90            & 65.55                     \\
MVRML   [2021]   & 61.44         & 51.16          & 74.72           & 77.72         & 66.26                 \\
FACT [2021]     & 60.34          & 54.85            & 74.48         & 76.55         & 66.56                 \\
TAF-Cal (Ours)          & \textbf{61.52}        & \textbf{54.98}             & \textbf{74.87}          & \textbf{76.97}           & \textbf{67.09}                 \\

\specialrule{.1em}{.1em}{0.1em} 
\end{tabular}
}
\caption{\label{table:office_comparison}Evaluation accuracy (\%) on OfficeHome with ResNet-18.}
\end{table}

\begin{table}[t!]
\resizebox{\columnwidth}{!}{%
\centering 
\footnotesize
\setlength{\tabcolsep}{4pt}
\renewcommand{\arraystretch}{1.10}
\begin{tabular}{l|ccc|c}
\specialrule{.1em}{.1em}{0.1em} 
\textbf{Method} & \textbf{Utrecht} & \textbf{Amsterdam} & \textbf{Singapore} & \textbf{Avg} \\ \hline
% \multicolumn{6}{c}{\textbf{ResNet-18}}                                                                                                   \\ \hline
DeepAll       & 0.183        & 0.619            & 0.781           & 0.528                                \\
BigAug [2020]          & 0.534        & 0.691             & 0.711          & 0.645                          \\
UDA [2020]     & 0.529          & \textbf{0.737}           & 0.782            & 0.683                          \\
MixDANN [2021] & 0.694          & 0.700                 & 0.839                 & 0.744                         \\
TAF-Cal (Ours)          & \textbf{0.736}        & 0.732             & \textbf{0.860}          & \textbf{0.776}                            \\

\specialrule{.1em}{.1em}{0.1em} 
\end{tabular}
}
\caption{\label{table:wmh_comparison}Evaluation DSC on WMH Challenge with U-Net.}
\end{table}

\begin{table}[t!]
\resizebox{\columnwidth}{!}{%
\centering 
\footnotesize
\setlength{\tabcolsep}{4pt}
\renewcommand{\arraystretch}{1.10}
\begin{tabular}{ccc|cccc|c}
\specialrule{.1em}{.1em}{0.1em} 
\textbf{AAF} & \textbf{TF-Cal (Train)} & \textbf{TF-Cal (Test)} & \textbf{Art} & \textbf{Cartoon} & \textbf{Photo} & \textbf{Sketch} & \textbf{Avg} \\ \hline
- & - & - & 77.95 & 77.92 & 95.68 & 69.56 & 80.28 \\
\checkmark & - & - & 85.35 & 78.63 & \textbf{96.23} & 77.58 & 84.45 \\
- & \checkmark & - & 80.76 & 77.22 & 96.18 & 69.33 & 80.88 \\
\checkmark & \checkmark & - & 84.60 & 78.88 & 95.23 & 77.17 & 83.97 \\
- & \checkmark & \checkmark & 83.31 & 81.23 & 95.63 & 77.45 & 84.41 \\
\checkmark & \checkmark & \checkmark & \textbf{85.70} & \textbf{82.55} & 96.07 & \textbf{82.60} & \textbf{86.73} \\
\hline\multicolumn{8}{c}{\textbf{TF-Cal with a Random Prototype}}                                                                                                    \\
\hline
- & \checkmark & \checkmark & 71.43 & 76.32 & 85.69 & 68.95 & 75.60 \\
\specialrule{.1em}{.1em}{0.1em} 
\end{tabular}

}
\caption{\label{table:pacs_ablation} Ablation Studies with ResNet-18 on PACS}
\vspace{-10pt}
\end{table}

\subsection{Further Analysis}
\subsubsection*{Ablation Studies of Different Components}
We investigate a thorough ablation studies of different components in our method. With only AAF, our method already outperforms most of the methods listed in Table \ref{table:pacs_comparison}. The AAF also achieves similar performance as FACT. As showed in \cite{Xu2021AFF}, FACT's augmentation component obtains accuracy of 83.44 which is lower than our AAF, which empirically verifies the merit of AAF in generating more diverse styles in features. To prove that using TF-Cal does not serve as a way for augmenting source data, we show that only applying TF-Cal at training does not have much difference in performance with the DeepAll. The similar observation is also found when both AAF and TF-Cal (Train) are activated. When we activate TF-Cal at both train-time and test-time, we see the great improvement, which also beats most of the methods in Table \ref{table:pacs_comparison}. Moreover, we see that AAF and TF-Cal can complement each other and further boost the generalizability. Finally, we empirically conduct the experiment on the importance of the source prototype. We randomly sample a prototype from Gaussian distribution for both TF-Cal (Train) and TF-Cal (Test), and we see a significant performance drop.

\subsubsection*{Visualization of Features}
\begin{figure}[h]
  \begin{subfigure}{0.495\columnwidth}
  \includegraphics[width=\textwidth]{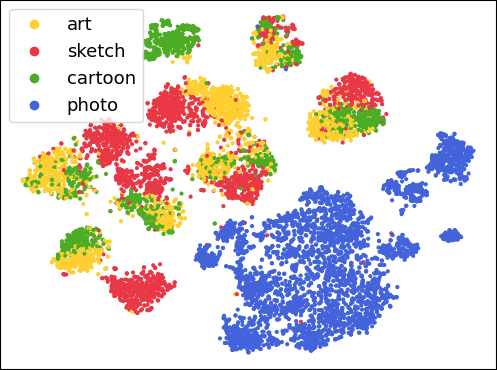}
  \caption{DeepAll}
  \end{subfigure}
  \hfill
  \begin{subfigure}{0.495\columnwidth}
  \includegraphics[width=\textwidth]{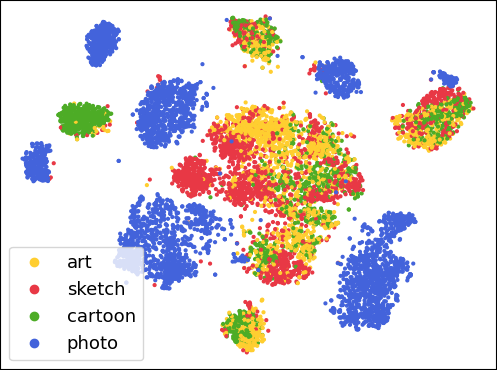}
  \caption{AAF} 
  \end{subfigure} 
  \begin{subfigure}{0.495\columnwidth}
  \includegraphics[width=\textwidth]{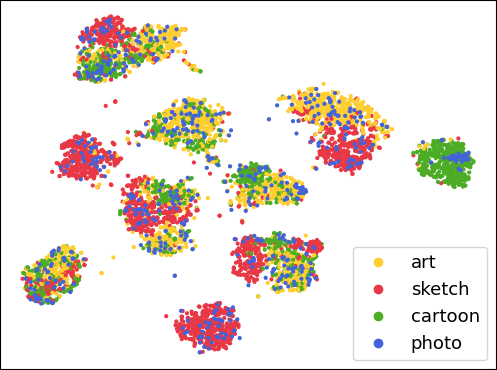}
  \caption{TF-Cal} 
  \end{subfigure} 
  \vspace{-5pt}
  \begin{subfigure}{0.495\columnwidth}
  \includegraphics[width=\textwidth]{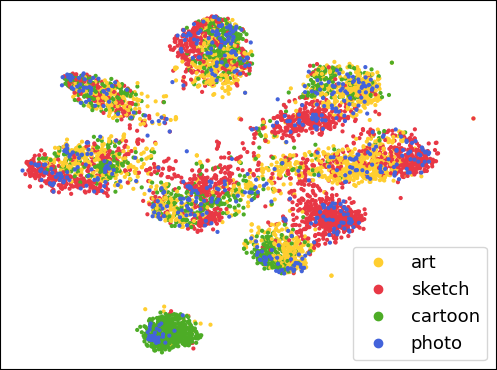}
  \caption{TAF-Cal} 
  \end{subfigure} 
  \caption{t-SNE visualization on PACS with sketch as target domain.}
  \label{fig:fig-tsne}
\end{figure}

To qualitatively analyze our contribution in calibrating style features, we applied t-SNE to compare the features learned by the DeepAll and our approach on PACS dataset with sketch as the target domain. From the Fig. \ref{fig:fig-tsne}, DeepAll has a satisfactory capacity for categorizing features, but it fails to align the source domain Photo and the target domain Sketch with other source domains. Specifically, AAF can well align the Sketch with the source domains, but with unclear cluster boundaries due to the mixing of source styles. TF-Cal precisely calibrates the target domain to the source domains, with distinct cluster boundaries. With the TF-Cal's advantage of aligning styles and AAF's advantage of diversifying styles, we observe a robust representation toward the target domain.

\subsubsection*{Hyperparameter Sensitivity}
\begin{figure}[h]
  \begin{subfigure}{0.495\columnwidth}
  \includegraphics[width=\textwidth]{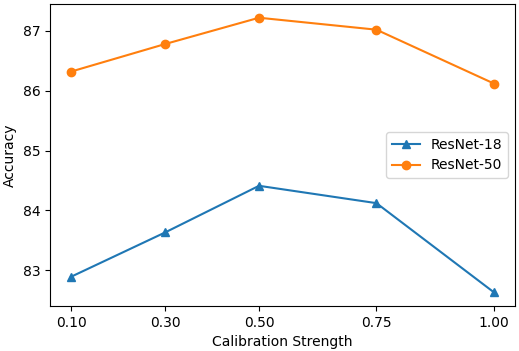}
  \caption{}
  \end{subfigure}
  \hfill
  \begin{subfigure}{0.495\columnwidth}
  \includegraphics[width=\textwidth]{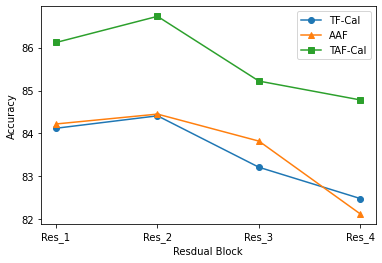}
  \caption{} 
  \end{subfigure} 
  \vspace{-8pt}
  \caption{(a) Hyperparameter Sensitivity on $\eta$ and $\tau$ for TF-Cal (b) Where to Apply Our Method.}
  \label{fig:fig-hyper}
\end{figure}

We conduct the hyperparameter experiment (Fig.\ref{fig:fig-hyper}a) on the calibration stength hyperparameter $\eta$ and $\tau$ for TF-Cal with ResNet-18 and ResNet-50 on PACS. As we define in the method section, $\eta$ and $\tau$ share the same value. We choose $\eta$ and $\tau$ from $\{0.1, 0.3, 0.5, 0.75, 1\}$. The larger $\eta$ and $\tau$ are, the more source prototype and the less target style in the target features. Both ResNet-18 and ResNet-50 have the similar trend. We find that TF-Cal performs the best when the $\eta$ and $\tau$ are equal to 0.5. When the $\eta$ and $\tau$ are smaller, the target style is less calibrated. When the $\eta$ and $\tau$ are larger, class-discriminative information in target styles might be reduced.

\subsubsection*{Where to Apply the Method}
We explore applying TF-Cal+AAF after different residule blocks of ResNet-18 on PACS. From the result in Fig.\ref{fig:fig-hyper}b, we see that applying our method after the second residual block performs the best, and the reason is that early layers extract more domain information than the later layers. As we can see that applying our method to the third and the fourth residual blocks, the performance drops. Because the these two blocks are closer to the classification layer, augmenting or calibrating the features will unnecessarily affect the class-discriminative features.

%% file: 6_conclusion.tex
\section{Conclusion}

In this paper, we propose a test-time calibration function for target styles extracted by Fourier transformation, and a feature style augmentation technique that complements style calibration. With TF-Cal, we alleviate the stress of having no knowledge about target domains during training by calibrating target styles with the source prototype at test-time. Additionally, AAF efficiently diversifies source amplitudes using higher-level style information extracted from features to help TF-Cal, outperforming image-based amplitudes augmentation. While our method TAF-Cal can be easily integrated into any CNN network, we achieve state-of-the-art performance on two primary tasks (i.e., classification and segmentation), and we conduct extensive experiments and qualitative analysis that validate our method's effectiveness and robustness.

%% file: ijcai22-multiauthor.bbl
\begin{thebibliography}{}

\bibitem[\protect\citeauthoryear{Albuquerque \bgroup \em et al.\egroup
  }{2019}]{Albuquerque2019GeneralizingTU}
Isabela Albuquerque, Jo{\~a}o Monteiro, Mohammad Javad~Darvishi Bayazi,
  Tiago~H. Falk, and Ioannis Mitliagkas.
\newblock Generalizing to unseen domains via distribution matching.
\newblock {\em arXiv: Learning}, 2019.

\bibitem[\protect\citeauthoryear{Balaji \bgroup \em et al.\egroup
  }{2018}]{Balaji2018MetaRegTD}
Yogesh Balaji, Swami Sankaranarayanan, and Rama Chellappa.
\newblock Metareg: Towards domain generalization using meta-regularization.
\newblock In {\em NeurIPS}, 2018.

\bibitem[\protect\citeauthoryear{Ben-David \bgroup \em et al.\egroup
  }{2009}]{BenDavid2009ATO}
Shai Ben-David, John Blitzer, Koby Crammer, Alex Kulesza, Fernando~C Pereira,
  and Jennifer~Wortman Vaughan.
\newblock A theory of learning from different domains.
\newblock {\em Machine Learning}, 79:151--175, 2009.

\bibitem[\protect\citeauthoryear{Cha \bgroup \em et al.\egroup
  }{2021}]{Cha2021DomainGN}
Junbum Cha, Han-Cheol Cho, Kyungjae Lee, Seunghyun Park, Yunsung Lee, and
  Sungrae Park.
\newblock Domain generalization needs stochastic weight averaging for
  robustness on domain shifts.
\newblock {\em ArXiv}, abs/2102.08604, 2021.

\bibitem[\protect\citeauthoryear{Dou \bgroup \em et al.\egroup
  }{2019}]{Dou2019DomainGV}
Qi~Dou, Daniel~Coelho de~Castro, Konstantinos Kamnitsas, and Ben Glocker.
\newblock Domain generalization via model-agnostic learning of semantic
  features.
\newblock In {\em NeurIPS}, 2019.

\bibitem[\protect\citeauthoryear{Ganin \bgroup \em et al.\egroup
  }{2016}]{Ganin2016DomainAdversarialTO}
Yaroslav Ganin, E.~Ustinova, Hana Ajakan, Pascal Germain, H.~Larochelle,
  François Laviolette, Mario Marchand, and Victor~S. Lempitsky.
\newblock Domain-adversarial training of neural networks.
\newblock In {\em J. Mach. Learn. Res.}, 2016.

\bibitem[\protect\citeauthoryear{Geirhos \bgroup \em et al.\egroup
  }{2019}]{Geirhos2019ImageNettrainedCA}
Robert Geirhos, Patricia Rubisch, Claudio Michaelis, Matthias Bethge, Felix
  Wichmann, and Wieland Brendel.
\newblock Imagenet-trained cnns are biased towards texture; increasing shape
  bias improves accuracy and robustness.
\newblock {\em ArXiv}, abs/1811.12231, 2019.

\bibitem[\protect\citeauthoryear{He \bgroup \em et al.\egroup
  }{2016}]{He2016DeepRL}
Kaiming He, X.~Zhang, Shaoqing Ren, and Jian Sun.
\newblock Deep residual learning for image recognition.
\newblock {\em 2016 IEEE Conference on Computer Vision and Pattern Recognition
  (CVPR)}, pages 770--778, 2016.

\bibitem[\protect\citeauthoryear{Huang \bgroup \em et al.\egroup
  }{2020}]{Huang2020SelfChallengingIC}
Zeyi Huang, Haohan Wang, Eric~P. Xing, and Dong Huang.
\newblock Self-challenging improves cross-domain generalization.
\newblock In {\em ECCV}, 2020.

\bibitem[\protect\citeauthoryear{Kuijf \bgroup \em et al.\egroup
  }{2019}]{Kuijf2019StandardizedAO}
Hugo~J. Kuijf, Adri{\`a} Casamitjana, D.~Louis Collins, M.~Dadar, Achilleas
  Georgiou, and et~al.
\newblock Standardized assessment of automatic segmentation of white matter
  hyperintensities and results of the wmh segmentation challenge.
\newblock {\em IEEE Transactions on Medical Imaging}, 38:2556--2568, 2019.

\bibitem[\protect\citeauthoryear{Li \bgroup \em et al.\egroup
  }{2017}]{Li2017DeeperBA}
Da~Li, Yongxin Yang, Yi-Zhe Song, and Timothy~M. Hospedales.
\newblock Deeper, broader and artier domain generalization.
\newblock {\em 2017 IEEE International Conference on Computer Vision (ICCV)},
  pages 5543--5551, 2017.

\bibitem[\protect\citeauthoryear{Li \bgroup \em et al.\egroup
  }{2018a}]{Li2018LearningTG}
Da~Li, Yongxin Yang, Yi-Zhe Song, and Timothy~M. Hospedales.
\newblock Learning to generalize: Meta-learning for domain generalization.
\newblock {\em ArXiv}, abs/1710.03463, 2018.

\bibitem[\protect\citeauthoryear{Li \bgroup \em et al.\egroup
  }{2018b}]{Li2018DomainGW}
Haoliang Li, Sinno~Jialin Pan, Shiqi Wang, and Alex~Chichung Kot.
\newblock Domain generalization with adversarial feature learning.
\newblock {\em 2018 IEEE/CVF Conference on Computer Vision and Pattern
  Recognition}, pages 5400--5409, 2018.

\bibitem[\protect\citeauthoryear{Li \bgroup \em et al.\egroup
  }{2020}]{Li2020DomainAM}
Hongwei Li, Timo Loehr, and et~al.
\newblock Domain adaptive medical image segmentation via adversarial learning
  of disease-specific spatial patterns.
\newblock {\em arXiv: Computer Vision and Pattern Recognition}, 2020.

\bibitem[\protect\citeauthoryear{Matsuura and
  Harada}{2020}]{Matsuura2020DomainGU}
Toshihiko Matsuura and Tatsuya Harada.
\newblock Domain generalization using a mixture of multiple latent domains.
\newblock {\em ArXiv}, abs/1911.07661, 2020.

\bibitem[\protect\citeauthoryear{Nussbaumer}{1982}]{Nussbaumer1982TheFF}
Henri~J. Nussbaumer.
\newblock The fast fourier transform.
\newblock 1982.

\bibitem[\protect\citeauthoryear{Piotrowski and
  Campbell}{1982}]{Piotrowski1982ADO}
Lisa Piotrowski and Fergus~William Campbell.
\newblock A demonstration of the visual importance and flexibility of
  spatial-frequency amplitude and phase.
\newblock {\em Perception}, 11:337 -- 346, 1982.

\bibitem[\protect\citeauthoryear{Ronneberger \bgroup \em et al.\egroup
  }{2015}]{Ronneberger2015UNetCN}
Olaf Ronneberger, Philipp Fischer, and Thomas Brox.
\newblock U-net: Convolutional networks for biomedical image segmentation.
\newblock In {\em MICCAI}, 2015.

\bibitem[\protect\citeauthoryear{Sicilia \bgroup \em et al.\egroup
  }{2021}]{Sicilia2021DomainAN}
Anthony Sicilia, Xingchen Zhao, and Seong~Jae Hwang.
\newblock Domain adversarial neural networks for domain generalization: When it
  works and how to improve.
\newblock {\em ArXiv}, abs/2102.03924, 2021.

\bibitem[\protect\citeauthoryear{Venkateswara \bgroup \em et al.\egroup
  }{2017}]{Venkateswara2017DeepHN}
Hemanth Venkateswara, Jose Eusebio, Shayok Chakraborty, and Sethuraman
  Panchanathan.
\newblock Deep hashing network for unsupervised domain adaptation.
\newblock {\em 2017 IEEE Conference on Computer Vision and Pattern Recognition
  (CVPR)}, pages 5385--5394, 2017.

\bibitem[\protect\citeauthoryear{Xu \bgroup \em et al.\egroup
  }{2021}]{Xu2021AFF}
Qinwei Xu, Ruipeng Zhang, Ya~Zhang, Yanfeng Wang, and Qi~Tian.
\newblock A fourier-based framework for domain generalization.
\newblock {\em 2021 IEEE/CVF Conference on Computer Vision and Pattern
  Recognition (CVPR)}, pages 14378--14387, 2021.

\bibitem[\protect\citeauthoryear{Zhang \bgroup \em et al.\egroup
  }{2020}]{Zhang2020GeneralizingDL}
Ling Zhang, Xiaosong Wang, and et~al.
\newblock Generalizing deep learning for medical image segmentation to unseen
  domains via deep stacked transformation.
\newblock {\em IEEE Transactions on Medical Imaging}, 39:2531--2540, 2020.

\bibitem[\protect\citeauthoryear{Zhang \bgroup \em et al.\egroup
  }{2021}]{Zhang2021MoreIB}
Jian Zhang, Lei Qi, Yinghuan Shi, and Yang Gao.
\newblock More is better: A novel multi-view framework for domain
  generalization.
\newblock {\em ArXiv}, abs/2112.12329, 2021.

\bibitem[\protect\citeauthoryear{Zhao \bgroup \em et al.\egroup
  }{2021}]{Zhao2021RobustWM}
Xingchen Zhao, Anthony Sicilia, and et~al.
\newblock Robust white matter hyperintensity segmentation on unseen domain.
\newblock {\em 2021 IEEE 18th International Symposium on Biomedical Imaging
  (ISBI)}, pages 1047--1051, 2021.

\bibitem[\protect\citeauthoryear{Zhou \bgroup \em et al.\egroup
  }{2020}]{Zhou2020DeepDI}
Kaiyang Zhou, Yongxin Yang, Timothy~M. Hospedales, and Tao Xiang.
\newblock Deep domain-adversarial image generation for domain generalisation.
\newblock {\em ArXiv}, abs/2003.06054, 2020.

\bibitem[\protect\citeauthoryear{Zhou \bgroup \em et al.\egroup
  }{2021}]{Zhou2021DomainGW}
Kaiyang Zhou, Yongxin Yang, Yu~Qiao, and Tao Xiang.
\newblock Domain generalization with mixstyle.
\newblock {\em ArXiv}, abs/2104.02008, 2021.

\end{thebibliography}
